\documentclass[runningheads]{llncs}
\usepackage{graphicx}
\usepackage{amsmath,amssymb} 
\usepackage{color}\usepackage[width=122mm,left=12mm,paperwidth=146mm,height=193mm,top=12mm,paperheight=217mm]{geometry}
\usepackage{enumitem}
\usepackage{caption}
\usepackage{subcaption}
\usepackage{xspace}
\usepackage{dsfont}
\newcommand{\abstractscenes}{ABSTRACT-50S\xspace}
\newcommand{\pascal}{PASCAL-50S\xspace}
\usepackage{cite}\usepackage[pagebackref=true,breaklinks=true,letterpaper=true,colorlinks,bookmarks=false,allcolors=green]{hyperref}
\usepackage{bm}

\begin{document}
\pagestyle{headings}
\mainmatter
\title{Collecting Image Description Datasets using Crowdsourcing} 

\titlerunning{Collecting Image Description Datasets using Crowdsourcing}

\authorrunning{Ramakrishna Vedantam, C. Lawrence Zitnick \and Devi Parikh}

\author{Ramakrishna Vedantam, C. Lawrence Zitnick \and Devi Parikh}
\institute{Virginia Tech\\ Microsoft Research}

\maketitle

\begin{abstract}
We describe our two new datasets with images described by humans. Both the datasets were collected using Amazon Mechanical Turk, a crowdsourcing platform. The two datasets contain significantly more descriptions per image than other existing datasets. One is based on a popular image description dataset called the UIUC Pascal Sentence Dataset, whereas the other is based on the Abstract Scenes dataset containing images made from clipart objects. In this paper we describe our interfaces, analyze some properties of and show example descriptions from our two datasets. 
\end{abstract}

\section{Introduction}
Recent works have explored the connection between Natural Language and Images. There is particular interest in both the Vision and NLP communities to explore the common ground between the two areas. On the vision side, understanding the interplay with language can help drive vision systems that communicate with humans, summarize important aspects in the scene etc. On the language side, there is much interest in grounding language learning with perceptual cues. 

To facilitate and spur future progress in these areas, appropriate datasets are critical. A lot of datasets of image descriptions exist~\cite{ZitnickCVPR2013,Rashtchian:2010:CIA:1866696.1866717,Ordonez:2011:im2text,journals/jair/HodoshYH13,Mller:2010:IEE:1869912}. However, the most number of sentences collected by any dataset so far is five. We introduce two image-description datasets with 50 captions for every image. We call these datasets \abstractscenes and \pascal. Our two datasets are ``gold-standard" in the sense that the sentences are all written by human subjects with the intention of describing the image. The \abstractscenes dataset is based on the dataset of Zitnick and Parikh~\cite{ZitnickCVPR2013} which has cartoon-like abstract images. This dataset, synthetically generated using crowdsourcing, provides opportunities to focus on image-semantics without the inhibition of (still) noisy visual detectors. The second dataset, is based on images from the UIUC Pascal Sentence Dataset. These are real images collected from Flickr. 

The UIUC Pascal Sentence Dataset has been used in various works for describing images~\cite{Kulkarni:2011:BTU:2191740.2191941,Mitchell:2012:MGD:2392712.2392740}, doing better semantic segmentation~\cite{conf/cvpr/FidlerSU13}, understanding properties of image descriptions~\cite{conf/cvpr/BergBDDGHMMSSY12}, etc. Other works have leveraged Abstract Images for performing zero-shot learning~\cite{Antol2014}, generating images from text~\cite{Zitnick_2013_ICCV}, and understanding the semantics of images~\cite{ZitnickCVPR2013}. We hope our new datasets will facilitate further progress along these varied directions.
\begin{figure}[tbp]
\centering
\includegraphics[width=\columnwidth]{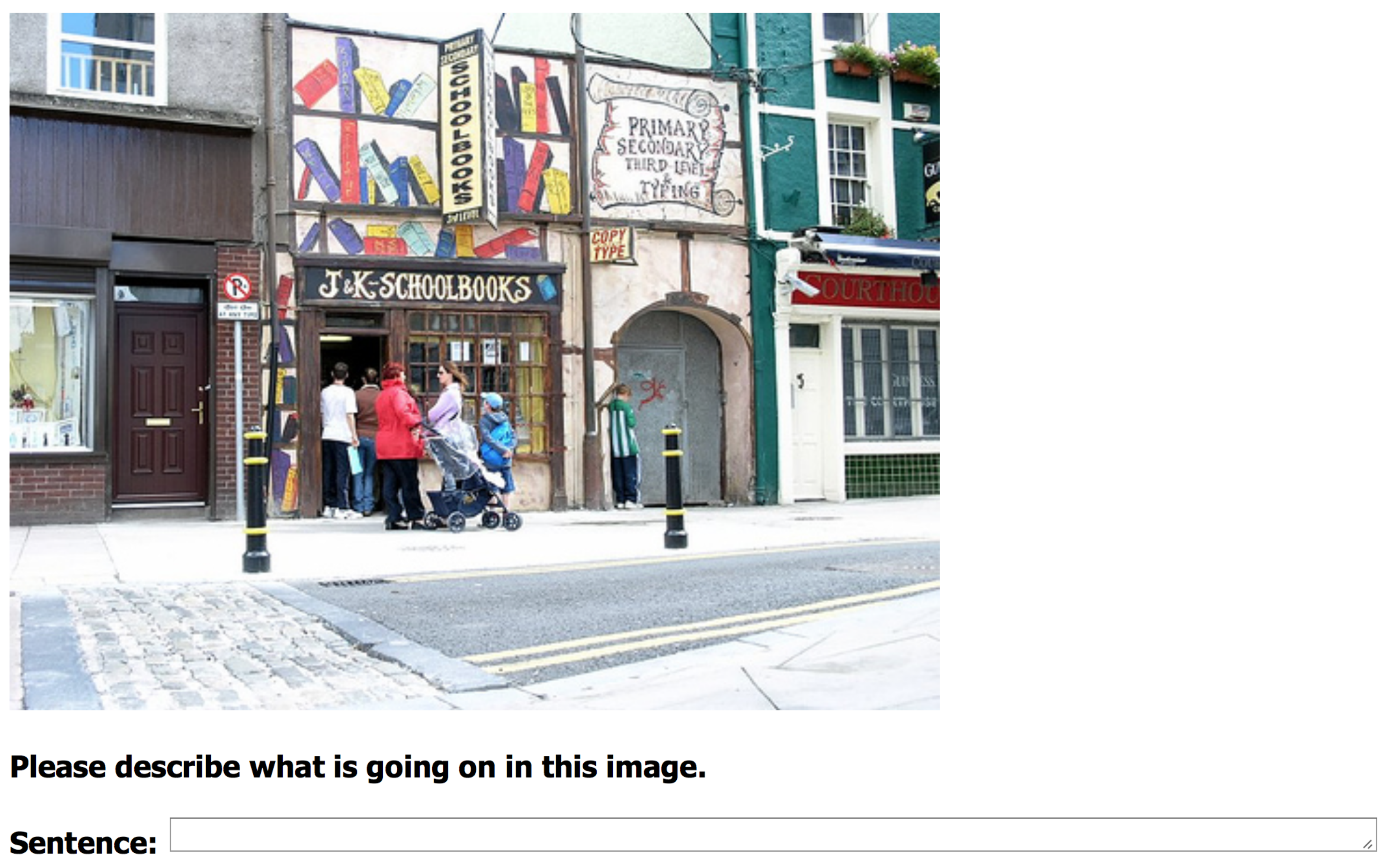}
\caption{A snapshot of our sentence collection interface, shown to subjects on Amazon Mechanical Turk.}
\end{figure}
\section{Related Work}
The most popular Image-Sentence dataset is the UIUC Pascal Sentence Dataset~\cite{Rashtchian:2010:CIA:1866696.1866717}. This dataset contains 5 human written descriptions for 1000 images. The SBU captioned photo dataset \cite{Ordonez:2011:im2text} contains one description per image for a million images, mined from the web. This dataset has automatically mined descriptions, which are not ``gold-standard". That is, there exist many descriptions which are not relevant to the image content~\cite{journals/jair/HodoshYH13}. Recent works have looked at the problem of identifying ``visual" text ~\cite{conf/acl/KuznetsovaOBBC13}. Further progress could lead to image description datasets with a large number of images. Our focus, in contrast, is to create a dataset that captures fine grained notions of object importance and description styles. Thus we create a dataset with a large number of descriptions per Image. The Abstract Scenes dataset contains cartoon like images with two descriptions. The recently released MS-COCO dataset~\cite{LMBHPRDZ:ECCV:2014} contains five sentences for a collection of over 100k images. The Flickr8k dataset contains five descriptions for a collection of 8000 images. The images in this dataset were queried for actions. Subjects were instructed to describe the major actions and objects in the scene. Other datasets of images and associated descriptions includes the Image Clef Dataset, which tends to have longer sentences (21 words)~\cite{Mller:2010:IEE:1869912}. We describe two new datasets. First is the \pascal dataset where we collect 50 sentences per image for the 1000 images from UIUC Pascal Sentence datset. The second is the \abstractscenes dataset where we collect 50 sentences for a subset of 500 images from the Abstract Scenes dataset. 
\section{Interface}
Our goal was to collect image descriptions that are objective and representative of the image content. We first showed subjects a set of images on mechanical turk and asked them to ``describe" them. We found that when asked to describe images, subjects would use their imagination and often not produce descriptions that are relevant to image content. We thus asked the subjects to ``transcribe" the major aspects of the scene into descriptions. We found that making this change helped elicit more objective and image-related descriptions from the Subjects.
\begin{figure}[tbp]
\centering
\begin{subfigure}{0.4\textwidth}
\includegraphics[width=\textwidth]{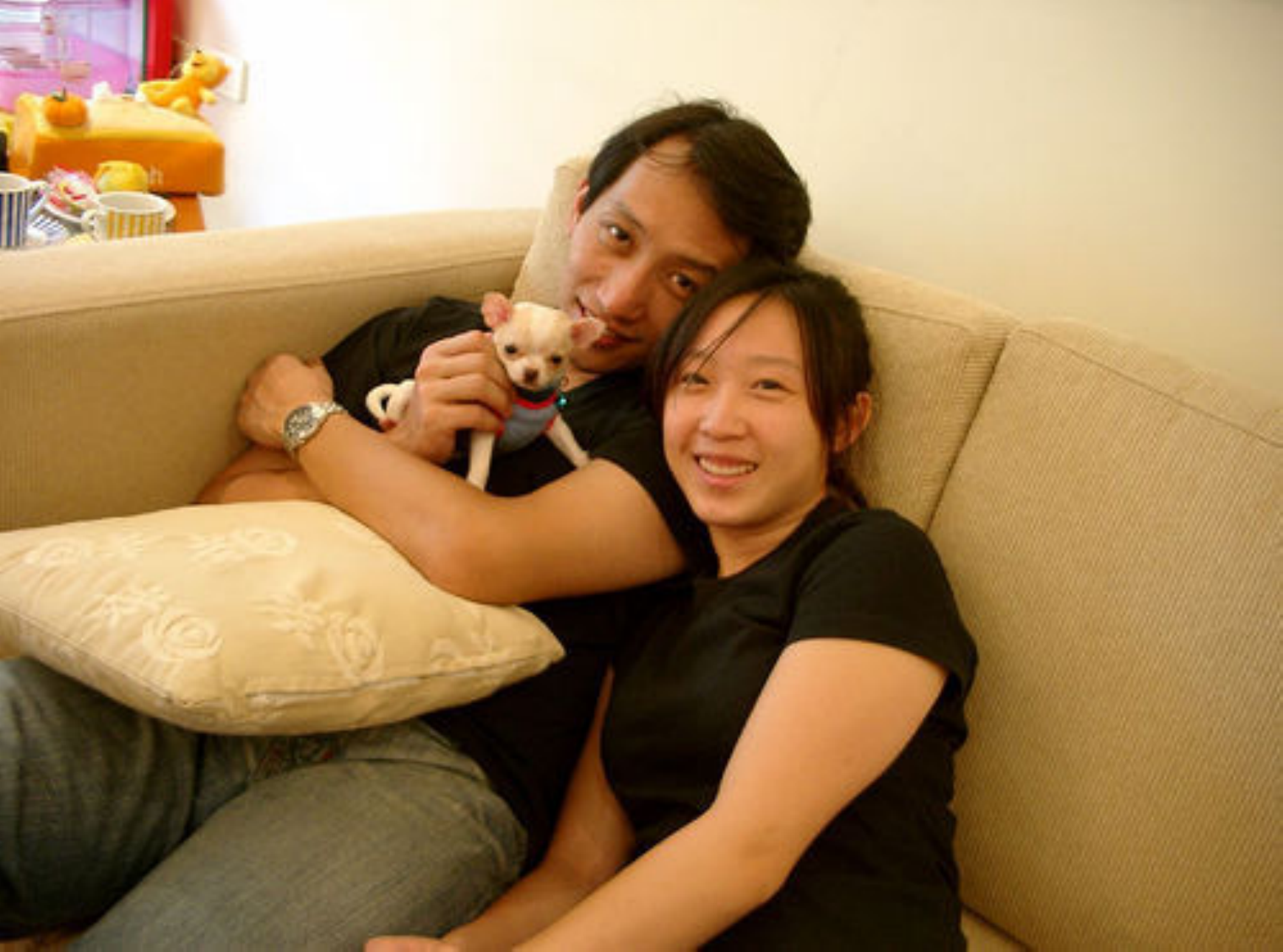}
\caption{A sample image shown to workers}
\end{subfigure}
\begin{subfigure}{0.4\textwidth}
\includegraphics[width=\textwidth]{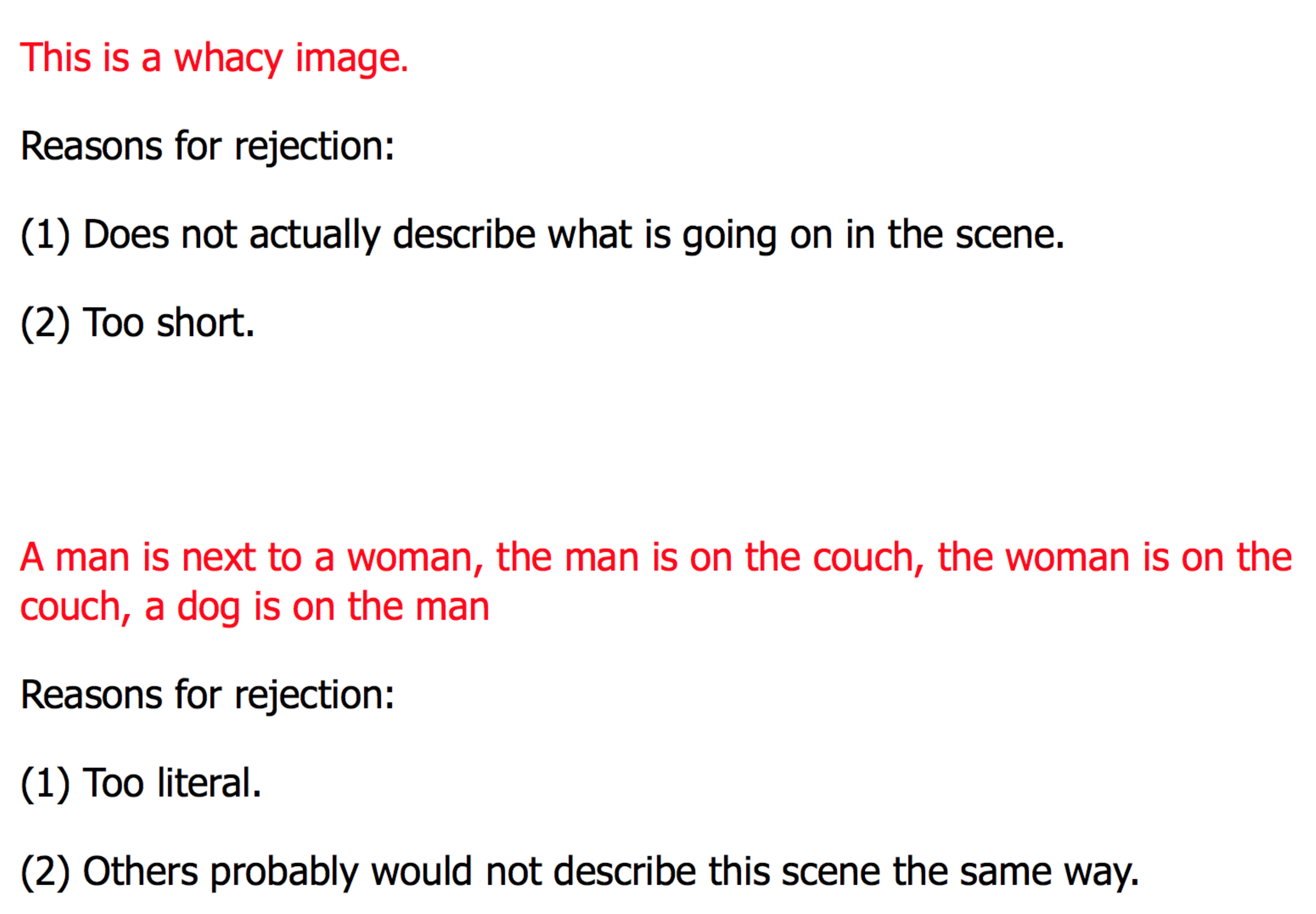}
\caption{Examples of sentences that would be rejected, with reasons}
\end{subfigure}
\caption{Explanation of our rejection criteria. Our goal was to collect sentences representative of image content}
\end{figure}
The exact details of our interface are as follows. Subjects are shown an image and a text box, asking them to describe what is ``going on" in the image. We instruct the workers that good transcriptions are those that others are also likely to provide (see Fig. 2). This includes writing descriptions rather than ``dialogs" or overly descriptive sentences. They were encouraged to capture the main aspects of the scene. Subjects were told that a good description should help others recognize the image from a collection of similar images. Instructions also mentioned that work with poor grammar would be rejected. Snapshots of our interface can be seen in Fig. 1. Overall, we had 465 subjects for \abstractscenes and 683 subjects for \pascal datasets. We ensure that each sentence for an image is written by a different subject.

To ensure that the sentences are of desired quality, certain qualification criteria were imposed. Subjects were required to be from the United States. Only subjects who had a 95\% HIT (Human Intelligence Task) approval rate and had been approved 500 times were considered eligible on Amazon Mechanical Turk. 

\section{Analysis and Results}
Overall, we find that the sentences collected for \pascal are on average 8.8 words in length. In contrast, on the \abstractscenes dataset we find that the description length is 10.59 words. This could be because of the tigher semantic sampling that the abstract images impose. Thus, sentences tend to be more detailed to be discriminative. We show some scenes from our \pascal dataset in Fig. 3 and some from our \abstractscenes dataset in Fig. 4 respectively. 
\begin{figure}
\includegraphics[width=\textwidth, page=1]{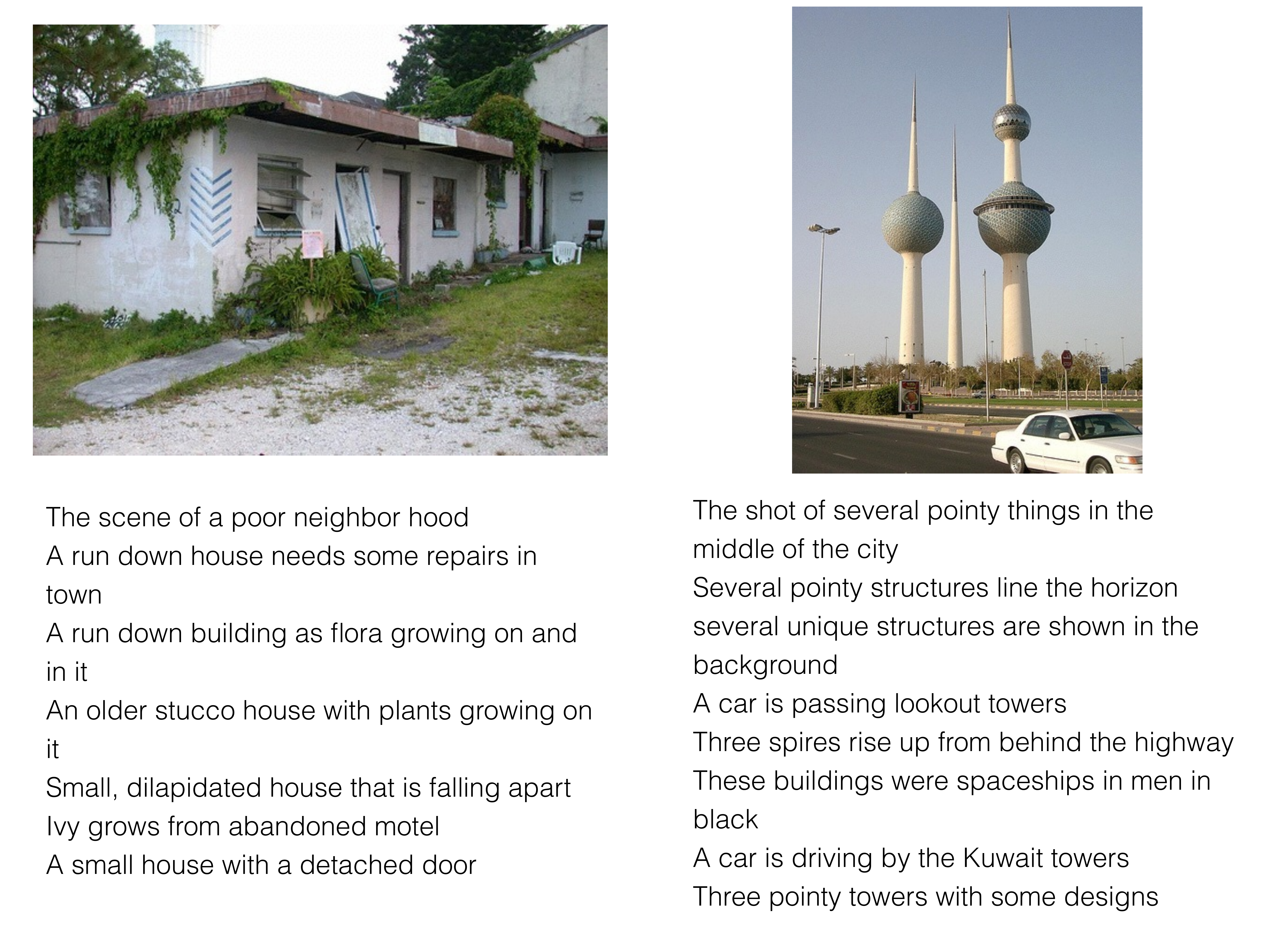}
\caption{Sample images with a subset of collected sentences from \pascal. Notice the rich variation in descriptions, because of collecting a large number of descriptions per image}
\end{figure}
\begin{figure}
\includegraphics[width=\textwidth, page=2]{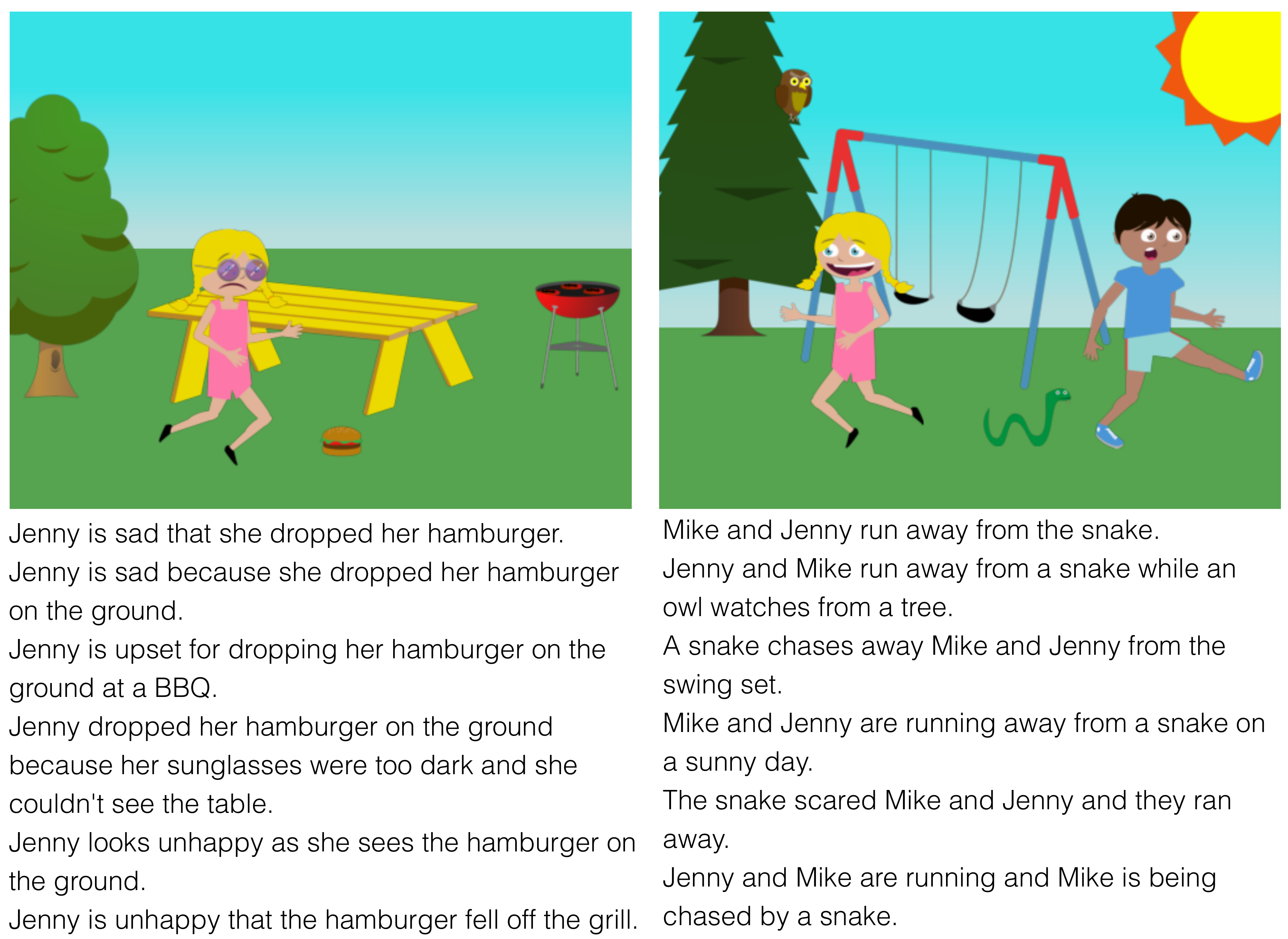}
\caption{Sample images with a subset of collected sentences from \abstractscenes. Notice the rich variation in descriptions, because of collecting a large number of descriptions per image}
\end{figure}
\section{Conclusions}
In this paper, we describe two new datasets \abstractscenes and \pascal with 50 sentences per image. We provide interface details and a background on the motivation for these datasets. These proposed datasets capture the many ways in which humans describe images. We hope these two new datasets will spur further research on exploring the connection between vision and language, two primary interaction modalities for humans, and lead to future research in building more intelligent systems.

\bibliographystyle{splncs}
\bibliography{collecting_descriptions}
\end{document}